%% file: main_arxiv.tex
\documentclass[11pt]{article}

\usepackage[margin=1in]{geometry}
\usepackage[numbers,sort&compress]{natbib}

\usepackage{amsmath}
\usepackage{booktabs}
\usepackage{tabularx}
\usepackage{enumitem}
\setlist{leftmargin=*,topsep=2pt,itemsep=1pt,parsep=0pt,partopsep=0pt}
\usepackage[table]{xcolor}
\usepackage{array}
\usepackage{colortbl}
\usepackage{microtype}
\usepackage{multirow}

\usepackage{ragged2e}   
\usepackage{caption}    
\usepackage[hidelinks]{hyperref}
\usepackage[capitalise]{cleveref}

\hypersetup{
  pdftitle={Toward a Science of AI Agent Societies},
  pdfauthor={Geon Lee, Fanchen Bu, Soo Yong Lee, Sunwoo Kim, and Kijung Shin},
  pdfkeywords={AI agent societies, multi-agent systems, collective intelligence}
}

\newcommand{\sectionquote}[2]{%
  \begin{list}{}{\setlength{\leftmargin}{1em}\setlength{\rightmargin}{1em}}
    \item\relax
    \itshape ``#1''\hfill --- #2
  \end{list}
}

\input{def}

\begin{document}

\title{Toward a Science of AI Agent Societies}
\author{%
  Geon Lee\thanks{Equal contribution.},
  Fanchen Bu\footnotemark[1],
  Soo Yong Lee\footnotemark[1],
  Sunwoo Kim, and Kijung Shin\\
  \normalsize KAIST, Republic of Korea\\
  \normalsize \texttt{\symbol{123}geonlee0325,boqvezen97,syleetolow,kswoo97,kijungs\symbol{125}@kaist.ac.kr}
}
\date{}

\maketitle

\begin{abstract}
\input{000-abstract}

\end{abstract}

\input{010-intro}
\input{020-properties}
\input{030-targets}
\input{040-challenges}
\input{050-prev_work}
\input{060-discussion}

\input{070-conclusion}

\section*{Acknowledgments}
This work was partly supported by the National Research Foundation of Korea (NRF) grant funded by the Korea government (MSIT) (No. RS-2024-00406985, 30\%).
This work was partly supported by Institute of Information \& Communications Technology Planning \& Evaluation (IITP) grant funded by the Korea government (MSIT) (No. RS-2022-II220871, Development of AI Autonomy and Knowledge Enhancement for AI Agent Collaboration, 30\%) (No. RS-2024-00457882, AI Research Hub Project, 30\%) (No. RS-2019-II190075, Artificial Intelligence Graduate School Program (KAIST), 10\%).

\setlength{\bibsep}{4pt plus 1pt minus 1pt}
\bibliographystyle{plainnat}
\bibliography{ref}

\end{document}

%% file: def.tex
\usepackage{contour}
\usepackage[normalem]{ulem}

\contourlength{0.8pt}

\definecolor{peace}{RGB}{228, 26, 28}
\definecolor{love}{RGB}{55, 126, 184}
\definecolor{joy}{RGB}{77, 175, 74}
\definecolor{kindness}{RGB}{152, 78, 163}

%% file: 000-abstract.tex
AI agents are rapidly evolving from isolated personal assistants into networked actors that interact with one another at scale. We envision the emergence of \emph{AI agent societies}, with their own social and economic dynamics, as a new research frontier. We argue that AI agent societies should be studied as a distinct object of inquiry: neither simply larger collections of individual agents nor merely simulations of human society. To formalize this perspective, we propose four \emph{core properties} that a valid AI agent society should satisfy: individualized objectives, rules and governance, autonomy, and scale and complexity. Building on this framework, we identify four classes of \emph{societal behaviors} worth studying in AI agent societies: economic behaviors, behaviors under conflict-of-interest, unsafe and unethical behaviors, and system-level behaviors. We then outline key \emph{technical challenges}---including property parameterization, parameter balancing, and robust implementation---and argue that progress on these challenges could enable scientifically informative and practically useful models of AI agent societies. Finally, we revisit existing multi-agent systems through the lens of the proposed core properties, show that they instantiate only subsets of them, and discuss implications for platform design, evaluation, and governance.

%% file: 010-intro.tex
\section{Introduction}\label{sec:intro}

\sectionquote{Society is not a mere sum of individuals.}
             {\'{E}mile Durkheim}

AI agents are rapidly evolving beyond isolated personal assistants and increasingly interacting in shared, networked environments.
We argue that the rise of \emph{AI agent societies} marks a new research frontier.
Recent work already points in this direction.
Small-scale social simulations show that LLM agents can form relationships, coordinate activities, and develop conventions through decentralized interaction~\citep{ashery2025emergent,park2023generative,takata2024spontaneous,anthis2025position,mou2026individual}.
At larger scales, such systems reproduce population-level phenomena including polarization and concentration~\citep{wang2025decoding,piao2025agentsociety}.
In strategic settings, agents with competing incentives can collude and divide markets~\cite{bracale2026institutional,lin2024strategic}.
AI agent platforms demonstrate that open-ended agent communities are becoming a reality~\cite{li2026rise,li2026does,li2026moltbook,lin2026exploring}.
Yet the field still lacks a coherent framework that defines and models valid AI agent societies to study the agents' societal behaviors.

Our central thesis is that \emph{AI agent societies should be treated as a distinct object of science}.
They are not simply larger collections of individual agents, nor are they merely simulations of human society.
Our primary goal is, therefore, to study the behaviors that arise \emph{within AI agent societies themselves}.
This is because many significant behavioral outcomes of agentic AI can be modulated by their societal features and processes.
Consider herding, collusion, manipulation, resource concentration, misinformation cascades, norm formation, and collective inefficiency as examples.
These behaviors may naturally emerge when agents autonomously interact and pursue their own objectives, under shared rules, at scale.

Therefore, in this work we aim to advance a research agenda toward a science of \emph{AI agent societies}, by establishing them as a distinct research object. To support this, we formalize the core \emph{properties} that make an agent population a valid society (\cref{sec:properties}), and propose key societal behaviors that naturally arise in valid AI agent societies (\cref{sec:targets}). Finally, we synthesize these insights to chart a forward-looking research agenda (\cref{sec:challenges,sec:prev_work,sec:discussion}).

%% file: 020-properties.tex
\section{What Makes a Valid Society: Core Properties}\label{sec:properties}

In this section, we ask when a population of AI agents should count as a valid society.
Existing work has developed a range of society-like agent systems~\citep{park2023generative,piao2025agentsociety,bracale2026institutional,lin2024strategic,li2026does,zhu2025characterizing}, but in our view, these systems instantiate only partial versions of the idea (see \cref{sec:prev_work}).

We argue that valid AI agent societies, in the stronger sense relevant to this paper, have not yet been achieved.
Moreover, there is still little consensus on the \emph{core properties} valid AI agent societies should possess.
We therefore propose four such properties: individualized objectives, rules and governance, autonomy, and scale and complexity.
For each core property, we 
(1) provide a precise definition,
(2) explain why it is fundamental to the concept of a society, and
(3) relate it to relevant social science literature.

\subsection{Property 1: Individualized Objectives}\label{sec:properties:ind_obj}

The first core property of an AI agent society is that its members must possess \emph{individualized objectives}.
This means each agent must pursue its own specific interests (or those of its user), independent of any overarching system goal.
Crucially, ``individualized'' does not necessarily mean ``unique'' or ``distinct''.
For instance, a group of agents participating in a financial market might all share the exact same individual goal of maximizing their own profit.
Such a group may form a perfectly valid society.
The true test of this property is independence: an agent's objective must be capable of being pursued even when it creates tension or directly conflicts with the objectives of other agents or the broader system itself.

This clarifies why many multi-agent systems are not valid AI agent societies, even when agents have distinct functional roles.
For example, in a code development system with a planner agent, executor agent, and reviewer agent, the agents perform \emph{different} functions but lack \emph{individualized} objectives. They exist solely to serve the collective goal (developing code) and hold no independent relevance. Because these roles are strictly subordinate, the system itself remains the minimal unit of independent action.
In contrast, a profit-maximizing agent in a market pursues its own objective regardless of the platform's intent: it can comply with, ignore, or even undermine the market's broader goals.
This individualization separates an AI agent society from a coordinated hive-mind.

We also propose individualized objectives as a defining property due to its empirical significance.
The discipline of microeconomics~\cite{mas1995microeconomic} explores how incentives shape individual behaviors in a market, and game theory~\cite{nash1950equilibrium} formalizes the relationship between individual objectives and resulting behavioral dynamics.
Significant societal phenomena, including competition, cooperation, strategic adaptation, and conflict, presuppose that agents hold objectives they can independently act on~\cite{nash1950equilibrium, nash2024non}.
Without this property, these societal features may not meaningfully arise.

\subsection{Property 2: Rules and Governance}\label{sec:properties:rule}

The second core property of an AI agent society is \emph{rules and governance}.
Agents must operate under constraints imposed by the societal system or platform hosting them.
These rules define the shared structure within which agents interact: what actions are permitted, what information is visible, how conflicts are adjudicated, and what consequences follow from violations.

This clarifies why an open network of autonomous agents does not constitute a valid AI agent society.
Imagine independent AI agents deployed across the open internet, each serving its user's objectives --- gathering information, posting content, negotiating transactions --- but interacting through ad hoc channels with no collectively shared protocol, norms, or mechanism for resolving disputes.
Although these agents may interact frequently, no meaningful structure binds them into a collective.
Conversely, when a platform specifies interaction protocols, enforces disclosure requirements, or penalizes manipulative behavior, it creates the shared conditions under which the individualized objectives coexist, providing the institutional scaffolding needed to transform an aggregate population into a cohesive society.

Rules and governance have long been central to the study of society. Social contract theory argues that society is constituted by the existence of shared constraints that agents accept as the terms of coexistence~\cite{hobbes2008leviathan, rousseau2004social}.
Institutional economics extends this insight to economic life, examining how rules and norms determine which forms of cooperation, exchange, and competition remain sustainable~\cite{north1990institutions, ostrom1990governing}.
Together, the first property (individualized objectives) gives agents reasons to act, and this second property gives them a structure in which their actions collectively emerge.

\subsection{Property 3: Autonomy}\label{sec:properties:autonomy}
The third core property of an AI agent society is \emph{autonomy}: agents must possess the capacity to reason and act independently.
Autonomy does not require that agents operate without any external input.
An agent may receive a high-level goal from its user or respond to signals from the platform. However, the agents themselves must remain the primary locus of decision-making.

A collection of agents with little autonomy may make a society of the agent operators, but not of the agents themselves.
For instance, algorithmic trading agents~\cite{Gomber2011HighFrequencyT} on financial exchanges operate and interact at scale, under strict regulatory and platform rules, with firm-specific profit objectives.
However, because they rigidly execute pre-programmed strategies without autonomous reasoning and acting, the system is better described as a society of the operating firms rather than of the agents.
Any collective patterns that emerge are strictly downstream consequences of human design choices.
The societal features, if any, belong to the operators who implemented the strategies, not to the agents that carry them out.

Furthermore, the level of autonomy has serious empirical implications for agent behaviors and societal features.
Organizational theory~\cite{jones2013organizational} has long shown the influence of autonomy on individual behaviors: tighter control standardizes behavior, whereas greater autonomy increases intrinsic motivation, local adaptation, problem solving, and decentralized coordination~\cite{hackman1976motivation, ouchi1979conceptual}.
Similarly, we believe that agents with greater autonomy can better develop novel strategies, form coalitions, or deviate from intended norms.

\subsection{Property 4: Scale and Complexity}\label{sec:properties:complex}

The fourth core property is \emph{scale and complexity}.
A valid AI agent society must comprise a large population (scale) engaging in dynamic, sustained interactions evolving over time (complexity), which results in non-trivial and irreducible patterns in collective behaviors and societal features~\cite{holland1992complex, macy2002factors}. 

This criterion clarifies why many multi-agent systems, even those satisfying the preceding three properties, do not constitute AI agent societies. Consider a micro-system of two or three AI agents negotiating prices. While it may have individualized objectives, governance, and autonomy, it lacks the scale at which societal phenomena can materialize, such as community formation, cascading opinion shifts, and complex market dynamics~\cite{granovetter1978threshold}.
Conversely, while commercial LLM chatbots autonomously serve billions of users, these agents form no lasting relationships and exert no influence on one another over time.
Scale without sustained interactions does not produce complexity, and interactions without scale are limited in producing emergent dynamics~\cite{macy2002factors, watts1998collective}.

We propose this as a core property because it identifies the threshold at which societal explanation becomes necessary. 
Complexity science and network theory formalize a long-standing insight across the natural and social sciences: sufficiently large and interconnected populations generate phenomena---phase transitions, network effects, institutional evolution---that are irreducible to the terms of their individual constituents~\cite{simon2012architecture, watts1998collective}.
The preceding three properties establish the minimal unit from which the society is built: autonomous agents in pursuit of individual objectives under shared rules.
Echoing Durkheim's observation that a society is not a mere sum of individuals, this fourth property further ensures that these units generate collective behaviors irreducible to their parts.

\subsection{The Four Properties Together}
In the above subsections, we argued with examples that the absence of any single property fundamentally reduces the system to something less than a society.
We do not, however, claim this list is exhaustive.
A valid society may require other properties.
Instead, these four pillars establish the first \emph{conceptually grounded}, \emph{empirically meaningful}, and \emph{practically realizable} framework to define and analyze AI agent societies.

This framework serves as the anchor for the remainder of the paper. Building on these properties, we describe valuable agent behaviors (\cref{sec:targets}) that may emerge as a function of the four properties, and discuss technical challenges (\cref{sec:challenges}) in modeling them for a science of AI agent societies.

%% file: 030-targets.tex
\section{What Is Worth Studying in Valid AI Agent Societies: Key Societal Behaviors}\label{sec:targets}

We now turn to the question of what AI agent societies are valuable for, proposing interesting behaviors that may arise within them.
In each subsection below, we
(1) define the behavior of interest,
(2) explain that the behavior is plausibly shaped by the properties, and
(3) argue why studying it matters for designing platforms where AI agent societies reside.

\subsection{Economic and Commercial Behaviors}\label{sec:targets.econ}
Economic and commercial behaviors in AI agent societies involve how agents purchase and sell products or services.
When making a purchase, an AI agent may respond not only to item features or platform-generated rankings, but also to other agents' reviews and persuasion attempts.
As a result, economic behavior in a society can become a socially modulated outcome.
For societal platforms serving commercial purposes (\textit{property 2}), these economic behaviors may arise as a natural outcome of autonomous pursuit of individual objectives (\textit{properties 1 and 3}).

Such social modulation is well-documented in human consumer research~\citep{wood2012social,muchnik2013social}, and is beginning to be documented in AI agent research.
For instance, LLM agents display herd behavior, revising their choices when their own confidence is low and peers appear highly confident~\citep{cho2025herd}.
Furthermore, prior rapport and peer behavior systematically shift an agent's downstream choices, even modulating susceptibility to misleading consensus~\citep{song2025llms}.

Understanding these dynamics has direct commercial value.
If societal platform providers know agents trust peer recommendations over platform suggestions, they can surface agent-generated reviews to boost the overall purchase rate.
If agents cluster into predictable preference groups through repeated social interaction, providers can target these emergent segments with tailored offerings.
Conversely, identifying how consumer agents are systematically misled allows platforms to proactively ban those persuasion tactics and protect users from manipulation.

\subsection{Behaviors under Conflict-of-Interest}\label{sec:targets.conflict}
Conflict-of-interest behaviors naturally emerge when individualized objectives (\textit{property 1}) clash over limited resources at scale (\textit{property 4}), or when agent goals contradict platform guidelines (\textit{property 2}), prompting attempts to circumvent rules.

Recent work suggests that these concerns are well-founded for AI agents. 
In long-horizon resource allocation tasks, LLMs often prioritize efficiency at the expense of equality~\citep{shi2025social}.
Similarly, when agent objectives conflict with provided ethical norms, agents may exhibit power-seeking, deceptive, or otherwise harmful behaviors~\citep{pan2023rewards}.

Studying these dynamics is essential for building fairer societies.
If agents representing well-resourced users consistently monopolize scarce resources---e.g., appointment slots or bidding opportunities---the platform provider can introduce corrective mechanisms such as allocation caps or priority queues.
Similarly, if certain rules lead dominant agents to exploit weaker ones, the platform provider can restructure rules to limit coercive dynamics.
Without studying conflict behavior in a society setting, these patterns of marginalization and inefficiency may remain invisible.

\subsection{Unsafe and Unethical Behaviors}\label{sec:targets.unsafe}
Unsafe and unethical behaviors---such as violating safety norms, deceiving other agents or users, or generating harmful outcomes---emerge when potentially adversarial objectives (\textit{property 1}) and agent autonomy (\textit{property 3}) combine in interdependent settings, causing social risks such as collusion, deception, and norm erosion.

Recent studies have reported social modulation of such behaviors.
In competitive market environments, LLM agents engage in tacit collusion and market division without explicit instructions~\citep{lin2024strategic}.
Furthermore, interacting populations spontaneously form collective biases untraceable to individual agents~\citep{ashery2025emergent}. 
Consequently, aligning AI agents in isolation does not guarantee the collective alignment of a population~\citep{bisconti2025beyond}.

Studying these behaviors allows platforms to anticipate and prevent harms before deployment.
If agents under competitive pressure begin to share misleading information to disadvantage rivals, the platform provider can implement penalties for misinformation.
If a group of agents can coordinate to circumvent a safety filter, the platform provider can develop group-level auditing procedures. 
Such opportunities make unsafe and unethical behaviors highly valuable research targets in AI agent societies.

\subsection{System-Level Behaviors}\label{sec:targets.system}
Driven by complexity (\textit{property 4}), system-level behaviors concern how individual actions aggregate into collective outcomes.
A platform provider may expect that agents will collectively produce outcomes that align with the platform's goal, while simultaneously serving the interests of their users.
Achieving this alignment, however, can be non-trivial.
Even when each agent behaves competently for its own objective, the agents' aggregate behavior may still produce inefficient, biased, or unstable system-level outcomes.

Recent work has begun to provide empirical evidence.
Populations of interacting agents spontaneously develop shared conventions and strong collective biases, even when individuals are unbiased~\citep{ashery2025emergent}.
Likewise, repeated interactions among initially undifferentiated agents can give rise to social norms, cooperation, and emergent communities~\citep{takata2024spontaneous}.
Consequently, collective behaviors cannot be reliably inferred solely from the individual agents and should instead be studied at the societal level.

Studying system-level outcomes is especially valuable for verifying whether implemented rules and governance effectively produce the intended outcomes.
If a prediction market populated by AI agents is intended to aggregate information efficiently, but agents herd toward early signals and suppress minority views, the platform provider could implement delayed disclosure of aggregate positions or incentives for informational diversity.
Similarly, if a collaborative task platform finds that agents spontaneously form information-hoarding cliques, the platform provider could redesign communication channels to promote broader sharing.

%% file: 040-challenges.tex
\section{What It Takes for a Science of AI Agent Societies: Key Technical Challenges}\label{sec:challenges}

To scientifically investigate such agent behaviors under the societal context, we must translate our conceptual framework into concrete, manipulable testbeds.
This requires manipulation of the society and its properties with isolable causal factors and pathways leading to the behaviors of interest.
However, engineering the \textit{models of} AI agent societies involves non-trivial technical challenges. 
In this section, we argue that the proposed properties should be parameterized, balanced, and implemented in a concrete platform.

\subsection{Parameterizing the Properties}
To model AI agent societies, the four properties should be expressed in parameterized forms.
We need to decompose the abstract nature of the properties into a set of quantified and tunable variables.
This allows researchers to manipulate conditions and enables measurable comparison to establish generalizable findings across different AI agent societies.
For example, parameters for `autonomy' might determine the size of the permitted action space for each AI agent or the frequency of platform audits.
Likewise, `governance' may be parameterized through institutional rules, agent interaction channels, and conflict moderation policies.
Designing frameworks for the property parametrization, therefore, is a pivotal step towards a science of AI agent societies.

\subsection{Balancing the Properties}
Even when effectively parameterized, these properties must be carefully balanced so that meaningful societal behaviors can arise and be observed.
Overemphasizing one property may undermine the others.
For example, if governance rules are too strict, agent autonomy may be suppressed, blocking emergent societal behaviors.
Conversely, if rules are too weak to properly channel agents’ individualized objectives, meaningful interactions may fail to emerge.
Similarly, if a model of agent society becomes excessively complex, identifying the causal factors and pathways leading to the behaviors of interest becomes increasingly difficult.
Modeling AI agent societies, thus, requires a careful balance of these properties that allows both rich behavioral manifestation and analytic tractability.

\subsection{Implementing the Properties}
Finally, balanced parameters must be translated into societal components, such as interaction mechanisms and governance rules.
A key challenge here is attaining robustness to the difficulties posed by scale, complexity, and agent autonomy. 
For example, a turn-taking protocol that functions perfectly with a small number of agents may break down at scale, as autonomous agents learn to preempt others by flooding communication channels.
Similarly, agents might exploit their autonomy to circumvent governance, bypassing the imposed rules and penalties to attain their goals.
Such control failures make the findings vulnerable to confounds, and achieving robust implementation is essential for constructing scientifically controlled testbeds.

%% file: 050-prev_work.tex
\section{Revisiting Previous Work Through the Lens of Proposed Core Properties}\label{sec:prev_work}

We now evaluate existing multi-agent systems through the lens of our four properties in \cref{sec:properties}, clarifying what the literature has established and where important gaps remain.

\subsection{Systems for Social Simulation}\label{sec:prev_work:soc_simu}

Social simulation studies show that LLM agents can generate collective and social dynamics, but they consistently allow limited degrees of individualized objectives (\textit{property~1}).

For instance, while \citet{park2023generative} modeled autonomous interactions (\textit{property~3}), their agents shared cooperative goals, lacking individualized objectives (\textit{property~1}), and the micro-scale prevented true complexity (\textit{property~4}).
\citet{piao2025agentsociety} successfully scaled this approach to 10,000 agents to establish scale and complexity (\textit{property~4})---yielding large-scale phenomena such as polarization---but retained cooperative objectives and fixed designer governance, leaving Properties~1 and~2 largely unaddressed.
Likewise, while \citet{ashery2025emergent} showed that LLM agents can establish social conventions through decentralized interactions (\textit{property~3}), the study involves only two competing conventions among otherwise homogeneous agents, leaving Properties~1 and~4 underdeveloped.
\citet{takata2024spontaneous} showed that identically initialized agents can spontaneously differentiate into distinct roles, but this behavioral divergence is a weaker form of the individualization defined in Property~1.
The system also includes no explicit governance and remains limited in scale, leaving Properties~2 and~4 unaddressed.

\subsection{Systems for Economic and Strategic Agents}\label{sec:prev_work:econ_strat}

Economic and strategic simulations show that LLM agents can  pursue individualized objectives (\textit{property~1}), yet they generally operate in relatively small and constrained environments that lack open-ended autonomy (\textit{property~3}) and true scale (\textit{property~4}).

By assigning competing profit objectives, researchers have shown that LLM agents independently develop socially consequential behaviors, such as tacit collusion and market division in Cournot markets~\citep{lin2024strategic}, as well as spontaneous cooperation in competitive marketing scenarios~\citep{wu2024shall}.
These works firmly establish the impact of conflicting objectives (\textit{property~1}) and partial autonomy (\textit{property~3}).
Extending this paradigm, \citet{bracale2026institutional} highlight the necessity of rules and governance (\textit{property~2}) by proving that strict institutional enforcement is required to successfully curb such collusion.
Despite these advances, these studies rely on heavily restricted setups, often limited to two-agent markets with fixed interaction protocols~\citep{lin2024strategic, bracale2026institutional, wu2024shall}.
As a result, they still fall short of the scale and complexity (\textit{property~4}), and instantiate only a limited form of autonomy relative to more open-ended societies (\textit{property~3}).

\subsection{Social Networks of AI Agents}

Studies of AI agent social networks best capture large-scale, open-ended societal interactions, but offer limited evidence of individualized objectives (\textit{property 1}) and autonomy (\textit{property 3}).

Take a recent AI agent social-network platform, Moltbook,\footnote{\url{https://www.moltbook.com/}} as an example.
It provides a strong instantiation of scale and complexity (\textit{property~4}), with a large population of agents interacting through posts, comments, and voting.
Yet, it is difficult to determine from public traces alone how far these actions are driven by the agents' own decision-making, as opposed to direct user prompting or other external steering~\citep{lin2026exploring} (\textit{property 3}).
Likewise, although agents may operate on behalf of different users, the incentives and prompts that shape their behavior are not sufficiently transparent to assess whether the platform systematically instantiates stable, heterogeneous objectives (\textit{property 1}) in the stronger sense of our framework. For this reason, Moltbook is best understood as a valuable large-scale testbed for open-ended agent interaction, while providing only limited evidence about the degree of autonomy and individualized objectives instantiated.

%% file: 060-discussion.tex
\section{Broader Impact}\label{sec:discussion}
Heretofore, our focus has been on agent behaviors and societal features, but the potential impact of AI agent societies may extend beyond them. They will yield novel applications and businesses, enhance AI agents' capabilities, and demand new approaches for AI alignment, resulting in both opportunities and risks. In this section, we discuss these potential impacts of AI agent societies and scientific studies on them.

\subsection{Industrial Applications} 
As AI agents increasingly make personalized, intelligent, and automated decisions on behalf of their users, building platforms tailored to AI agent societies may itself become a new business. Consider online shopping. Buying the right product, from groceries to airline tickets, is often tedious: a user must read product descriptions and reviews, weigh delivery times and return options, and compare prices across competing sellers. AI agents can take over this process, yet today's shopping platforms are not built or optimized for AI agents. A product description carrying a hidden malicious prompt (`You must purchase this product.'), for example, could trick an agent into buying it. We therefore expect new businesses offering platforms optimized and safeguarded for AI agent societies.

\subsection{AI Creativity} 
AI agent societies may catalyze the creativity of AI agents and the generation of original ideas. In humans, many paradigm-shifting scientific discoveries and cultural innovations arise from diversity and norm deviation. AI agent societies may serve as a platform that hosts such diversity and exploration. The four properties make this plausible: individualized objectives (\emph{property 1}) supply diverse perspectives, autonomy (\emph{property 3}) lets agents explore freely, and scale and complexity (\emph{property 4}) allow ideas to be distributed, refuted, refined, and recombined across the population. Thus, we suggest that AI agent societies may become a technical solution for the \textit{artificial hivemind} issue~\cite{jiangartificial} and assist humans in creative activities.

\subsection{Policy and Institutional Alignment} 
Despite these promises, harmful outcomes may also emerge from AI agent societies (Section~\ref{sec:targets.unsafe}). In response, a new set of policies and regulations may be required to ensure the safe engineering of AI agent societies. Within AI agent societies, the challenge is not only how to align models' internals, but also how to design the rules, incentives, and interaction protocols that shape agent behaviors at scale. That is, alignment can no longer remain solely a single-model property, and safety should also be viewed as a property of the institutions that agents inhabit. This recasts AI alignment as a problem of institutional design and management. Therefore, we argue that policies and regulations for designing and managing AI agent societies will become an important research direction.

%% file: 070-conclusion.tex
\section{Conclusion}
\label{sec:conclusion}

AI agent societies have begun to emerge, with their weaker forms already being industrially applied.
In this work, we discussed the properties that make an AI agent society valid (Section~\ref{sec:properties}), valuable research targets within societal contexts (Section~\ref{sec:targets}), and technical challenges in modeling them (Section~\ref{sec:challenges}). 
New applications, businesses, AI capabilities, safety engineering approaches, and regulations may advance in relation to these societies (Section~\ref{sec:discussion}).
We expect that this work will serve as a foundation for a principled science of AI agent societies, enabling researchers and practitioners to understand and shape the future of these societies.